\documentclass{article}

\usepackage[T1]{fontenc}
\usepackage{tgtermes}
\usepackage{amsmath}
\usepackage{amssymb}
\usepackage{amsfonts}

\usepackage{tgtermes}
\usepackage{amsmath}
\usepackage{scalefnt,letltxmacro}
\LetLtxMacro{\oldtextsc}{\textsc}
\renewcommand{\textsc}[1]{\oldtextsc{\scalefont{1.10}#1}}
\usepackage[scaled=0.92]{PTSans}

\usepackage[usenames,dvipsnames]{xcolor}
\definecolor{shadecolor}{gray}{0.9}

\usepackage[english]{babel}
\usepackage[parfill]{parskip}
\usepackage{afterpage}
\usepackage{framed}
\usepackage{nicefrac}
\usepackage{bm}
\usepackage{fixmath}

\usepackage[colorinlistoftodos,
           prependcaption,
           textsize=small,
           backgroundcolor=yellow,
           linecolor=lightgray,
           bordercolor=lightgray]{todonotes}

\usepackage{lineno}

\usepackage{ragged2e}

\DeclareRobustCommand{\parhead}[1]{\textbf{#1}~}

\usepackage{graphicx}
\usepackage[labelfont=bf]{caption}
\usepackage[format=hang]{subcaption}

\usepackage{booktabs}

\usepackage{natbib}

\usepackage{listings}
\usepackage{fancyvrb}
\fvset{fontsize=\normalsize}

\usepackage[colorlinks,linktoc=all]{hyperref}
\usepackage[all]{hypcap}
\hypersetup{citecolor=Violet}
\hypersetup{linkcolor=black}
\hypersetup{urlcolor=MidnightBlue}

\usepackage[nameinlink]{cleveref}
\creflabelformat{equation}{#1#2#3}
\crefname{equation}{eq.}{eqs.}  
\Crefname{equation}{Eq.}{Eqs.}

\usepackage[acronym,smallcaps,nowarn]{glossaries}

\usepackage{listings}
\lstdefinestyle{alp_style}{
    commentstyle=\color{OliveGreen},
    numberstyle=\tiny\color{black!60},
    stringstyle=\color{BrickRed},
    basicstyle=\ttfamily\scriptsize,
    breakatwhitespace=false,
    breaklines=true,
    captionpos=b,
    keepspaces=true,
    numbers=none,
    numbersep=5pt,
    showspaces=false,
    showstringspaces=false,
    showtabs=false,
    tabsize=2
}
\newacronym{ADVI}{advi}{automatic differentiation variational inference}

\newacronym{BBVI}{bbvi}{black-box variational inference}

\newacronym{CDF}{cdf}{cumulative density function}
\newacronym{CS-EFE}{cs-efe}{context selection for exponential family embeddings}
\newacronym{CTM}{ctm}{correlated topic model}

\newacronym[\glslongpluralkey={deep exponential families}]{DEF}{def}{deep exponential family}
\newacronym{DMIS}{dmis}{deterministic multiple importance sampling}

\newacronym{EFE}{efe}{exponential family embeddings}
\newacronym{ELBO}{elbo}{evidence lower bound}
\newacronym{EM}{em}{expectation maximization}

\newacronym{GNTS}{gn-ts}{gamma-normal time series model}
\newacronym{G-REP}{g-rep}{generalized reparameterization}

\newacronym{KL}{kl}{{K}ullback-{L}eibler}

\newacronym{LDA}{lda}{latent {D}irichlet allocation}

\newacronym{MF}{mf}{matrix factorization}
\newacronym{MIS}{mis}{multiple importance sampling}

\newacronym{OBBVI}{o-bbvi}{overdispersed black-box variational inference}

\newacronym{SVI}{svi}{stochastic variational inference}

\newacronym{VI}{vi}{variational inference}

\DeclareRobustCommand{\E}[2]{\mathbb{E}_{#1}\left[#2\right]}
\DeclareRobustCommand{\entropy}[1]{\mathbb{H}\left[#1\right]}

\DeclareRobustCommand{\tr}[1]{\textrm{tr}\left(#1\right)}

\newcommand{\dist}{\textsc{dist}}
\newcommand{\gp}{\textsc{gp}}

\newcommand{\Ncal}{\mathcal{N}}
\newcommand{\Rcal}{\mathcal{R}}

\newcommand{\bzero}{\mathbf{0}}

\newcommand{\bA}{\mathbf{A}}
\newcommand{\bb}{\mathbf{b}}

\newcommand{\bD}{\mathbf{D}}

\newcommand{\boldf}{\mathbf{f}}

\newcommand{\bH}{\mathbf{H}}

\newcommand{\bL}{\mathbf{L}}

\newcommand{\bR}{\mathbf{R}}

\newcommand{\bV}{\mathbf{V}}

\newcommand{\bW}{\mathbf{W}}
\newcommand{\by}{\mathbf{y}}

\newcommand{\bx}{\mathbf{x}}
\newcommand{\bX}{\mathbf{X}}

\newcommand{\bepsilon}{\mathbold{\epsilon}}

\newcommand{\bmu}{\mathbold{\mu}}

\newcommand{\bSigma}{\mathbold{\Sigma}}

\newcommand*{\QEDB}{\hfill\ensuremath{\square}}

\usepackage[accepted]{icml2018}

\begin{document}

\twocolumn[
\icmltitle{Non-Parametric Variational Inference with Graph Convolutional Networks for Gaussian Processes}




\begin{icmlauthorlist}
\icmlauthor{Linfeng Liu}{tufts}
\icmlauthor{Liping Liu}{tufts}
\end{icmlauthorlist}

\icmlaffiliation{tufts}{Department of Computer Science, Tufts University, Medford, MA}

\icmlcorrespondingauthor{Linfeng Liu}{linfeng.liu@tufts.edu}
\icmlcorrespondingauthor{Liping Liu}{liping.liu@tufts.edu}

\icmlkeywords{Machine Learning, ICML}

\vskip 0.3in
]



\printAffiliationsAndNotice{} 

\begin{abstract}
Inference for GP models with non-Gaussian noises is computationally expensive when dealing with large datasets. Many recent inference methods approximate the posterior distribution with a simpler distribution defined on a small number of {\it inducing points}. The inference is accurate only when data points have strong correlation with these inducing points. 
In this paper, we consider the inference problem in a different direction: GP function values in the posterior are mostly correlated in short distance. We construct a variational distribution such that the inference for a data point considers only its neighborhood. With this construction, the variational lower bound is highly decomposible, hence we can run stochastic optimization with very small batches. We then train Graph Convolutional Networks as a reusable model to identify variational parameters for each data point. Model reuse greatly reduces the number of parameters and the number of iterations needed in optimization. The proposed method significantly speeds up the inference and often gets more accurate results than previous methods.
 
\end{abstract}

\section{Introduction}
\label{introduction}
As a non-parametric Bayesian method, Gaussian Process (GP) \citep{rasmussen06} provides flexible and
expressive models for various machine learning tasks, such as regression, classification, and Bayesian optimization. The success of a GP model hinges on its efficient inference, which is especially true when the data is large in scale.

The central problem of GP inference is to calculate the posterior distribution of the function values at input points. Though the posterior of GP with non-Gaussian noise is often not Gaussian, many inference methods nevertheless approximate the posterior with a multivariate Gaussian distribution. Variational inference \citep{jordan99, wainwright08} does so by  maximizing the Evidence Lower Bound (ELBO) of the data likelihood. Standard 
variational inference for GP on large datasets is prehibitively expensive, as it needs to optimize the large covariance matrix of the variational distribution and also to inverse the prior covariance in gradient calculation. It is clear that we need further approximations to speed up the inference procedure. 

Variational inference based on inducing points \citep{quinonero05, titsias09} uses a low-rank matrix as the covariance of the variational distribution. It first defines a multivariate Gaussian distribution for function values at a small set of $M$ inducing points and then derives distributions of all other data points from the covariance defined by the prior. This method needs space $O(NM)$ and time $O(NM^2)$ in one gradient calculation, with $N$ being the number of data points. With stochastic optimization \citep{hoffman13},  GP inference \citep{hensman15, nguyen14, sheth15,dezfouli15, krauth16, cheng17} can be done in small batches, but each batch still needs to consider all inducing points. In terms of accuracy, inducing-point method has a requirement that all data points need to strongly correlate to a small number of inducing points, which may not be true in many applications. Methods in 
this category also lose the non-parametric flavor of GP, as the flexibility of the variational distribution is solely decided by the posterior distribution on inducing points.

We consider the approximation in an alternative direction. We propose to  1) localize the inference for a single data point within a realatively small neighborhood around it,  and 2) reuse the local inference procedure for all data points. With most popular kernel settings (Chapter 4 of \citep{rasmussen06}), the GP function value at a location is only impacted by function values at nearby locations in the prior, so is in the posterior. As a local neighborhood provides most information for the inference at a data point, neglect of weak long-distance correlation brings drastic computational advantage. 

In this work, we construct a variational distribution with which the inference for a data point takes the information only from its $K$ surrounding neighbors.
The variational distribution supports fast sampling from its marginals, then it works for GP with general noise distributions \citep{nguyen14, sheth15,dezfouli15, krauth16} with the reparameterization technique \citep{kingma13}. 
This construction makes the ELBO highly decomposible, so it is very cheap to get an unbiased estimation of the ELBO and run stochastic optimization with batch training. The update with a batch of size $N_b$ only takes time $O(N_bK^2)$.

Inspired by recognition networks used in variational inference \citep{kingma13, mnih2014, miao2016}, we further train two Graph Convolutional Networks (GCN) \citep{kipf2017} as the local inference procedure to identify variational parameters for each data point. The two networks take surrounding observations as the input and output variational parameters. The inference procedure reuses the two networks at all data points. This technique greatly reduces the number of variational parameters and optimization iterations in the entire inference procedure. 

The proposed inference methods (with and without an inference network) are non-parametric in the sense that the flexibility of the variational distribution grows with the data size. Especially, the method with an inference network does non-parametric inference with a parametric model. 

We compare our methods with other state-of-the-art methods on several tasks. The results show that our method achieves better inference performance while using much less computation time.



\section{Gaussian Process and Variational Inference}
\parhead{Gaussian Proces:}
A Gaussian process defines a distribution over the function space. If a function $f$ is from a Gaussian process
$\gp(m(\cdot), \kappa(\cdot, \cdot))$, then it implies that $\E{}{f(\bx)} = m(\bx)$ and 
$\mathrm{Cov}(f(\bx), f(\bx'))=\kappa(\bx, \bx')$, $\forall \, \bx, \bx' \in \Rcal^d$. 
Suppose we have $N$ data points, $\bX = (\bx_i: i=1, \ldots, N)$, $\by=(y_i: i=1, \ldots, N)$, and we 
want to model the conditional probability $p(y | \bx)$ with GP, then we can specify the conditional distribution as follows.
\begin{align}
f \sim \gp(0(\cdot), \kappa(\cdot, \cdot)), ~ 
\theta = \lambda(f(\bx_i)), ~ 
y_i \sim \dist(\theta). \label{gp}
\end{align}
The link function $\lambda(\cdot)$ and the distribution $\dist$ can be arbitrary as long as we can calculate $\log p(y_i | f(\bx_i))$ and take its gradient with respect to $f(\bx_i)$. This formulation is general enough for various regression and classification problems. 

Denote $\boldf = (f_i = f(\bx_i): i=1, \ldots, N)$, then $\boldf$ is a sample of Gaussian distribution with mean $\bzero$ 
and covariance $\bSigma = \kappa(\bX, \bX)$, the latter of which is referred as kernel matrix. 
GP inference concerns the calculation of the posterior distribution $p(\boldf | \bx, \by)$.
The inference problem is often intractable when the noise distribution is non-Gaussian, then approximate inference becomes inevitable. In this paper we focus on variational inference and approximate $p(\boldf | \bx, \by)$ 
with a variational distribution $q(\boldf)$. 

Assume now we can calculate the posterior $p(\boldf | \bx, \by)$ or its approximation $q(\boldf)$, we make prediction 
for new data points as follows. Let $\bx_{\star}$ be a new data point, then we calculate the predictive distribution as follows. 
\begin{align}
p(y_{\star} | \bx_{\star}, \bX, \by) &= 
\int_{f_{\star}} p(y_{\star} | f_{\star}) p(f_{\star} |\bx_{\star}, \bX, \by) ~ \mathrm{d} f_{\star} \nonumber\\
&\approx \int_{f_{\star}} p(y_{\star} | f_{\star}) q(f_{\star} |\bx_{\star}, \bX, \by) \mathrm{d} f_{\star}~. \label{prediction} 
\end{align}
Here $q(f_{\star} | \bx_{\star}, \bX, \by) = \int_{\boldf} p(f_{\star} | \boldf)q(\boldf) \mathrm{d}\boldf$, which is 
easy to calculate when $q(\boldf)$ is Gaussian. The integral in \eqref{prediction} may not have a 
closed-form solution, but we can easily approximate it with Monte Carlo samples, as the integral variable is in only one dimension.  

\parhead{Variational Inference for GP:} 
Variational inference searches for a good approximation $q(\boldf)$ by maximizing the ELBO, $L(q(\boldf))$, in \eqref{elbo} with respect to $q(\boldf)$.
\begin{multline}
 \log p(\by | \bX) \ge L(q(\boldf)) := \E{q(\boldf)}{\log p(\boldf) } \\
 + \E{q(\boldf)}{\log p(\by | \boldf)} + \entropy{q}. \label{elbo}
\end{multline}
The bound is tight when $q(\boldf)=p(\boldf | \bX, \by)$. For computational convenience, 
$q(\boldf)$ is often chosen to be a multivariate Gaussian distribution.
Denote its mean as  $\bmu$ and variance as $\bV$, then the optimization objective $L(q(\boldf))$ becomes $L(\bmu, \bV)$.

Direct calculation of \eqref{elbo} with a general GP model has three issues in real applications. 
The first term on the right side of \eqref{elbo} needs the inverse of prior covariance $\bSigma$. The second term often uses Monte Carlo samples to approximately the expectation $\E{q(f_i)}{p(y_i | f_i)}$, so it needs cheap samples from the marginal distribution $q(f_i)$. Third, the entropy term $\entropy{q}$ needs the determinant of $\bV$.   

Even one can calculate the ELBO in \eqref{elbo}, the optimization problem is still challenging for large problems, as there are a large number of optimization variables in $\bmu$ and $\bV$.


\section{Non-Parametric Variance Inference for general GP models}
\subsection{The variational distribution}

We construct the variational distribution $q(\boldf)$ as a multivariate Gaussian distribution with mean $\bmu$ and covariance $\bV$. The covariance $\bV$ is further parameterized as $ \bV= \bL\bL^\top$, with $\bL$ being a sparse lower triangular matrix. The off-diagonal non-zero elements in each row $\bL_i, 1 \le i \le N$ are indicated by a set $\alpha(i) \subseteq \{1, \ldots, i - 1\}$. The set $\alpha(i)$ has at most $K$ non-zero elements, $|\alpha(i)| = \min(K, i - 1)$. The sparse pattern of $\bL$ is,
\begin{align}
\bL_{ij} = 0 \quad \mathrm{if} ~ j \notin \alpha(i) \mbox{ or } j > i. 
\end{align}
The number of free variables in $\bL$ is $N(K + 1) - K(K+1)/2$. The second term $K(K+1)/2$ corresponds to the extra number of zeros for the first $K$ rows. The size requirement of $\alpha(i)$-s allows storing $\bL$ into a $N \times (K + 1)$ matrix, which is convenient for matrix computation in implementation.

With this construction, it is easy to sample from $q(\boldf)$. Let $\bepsilon \sim \Ncal(\mathbf{0}, \mathbf{I})$ be an $N$-dimensional 
Gaussian random white noise, then a sample $\boldf$ from $q(\boldf)$ can be generated by 
\begin{align}
\boldf = \bmu +  \bL\bepsilon.
\end{align}
It is fast to draw samples from the marginal $q(f_i)$: we only need to sample a $(K + 1)$-dimensional white noise $(\bepsilon_{\alpha(i)}, \epsilon_{i})$ and then calculate $f_i = \mu_i + \bL_{i,\alpha(i)} \bepsilon_{\alpha(i)} + \bL_{i,i} \epsilon_{i} $ as a sample. 

This construction of $q(\boldf)$ corresponds to a directed graphical model defined on all $N$ data points \citep{rue2005}, though a directed graphical model does not directly give marginal distributions. In the graph, $\alpha(i)$ is the parent set of data point $i$. We define $\alpha(\cdot)$ sets from the prior kernel matrix. The principle is that, if two data points $i$ and $j$, $j < i$, are strongly correlated in the prior, then $j$ is very likely to be in $\alpha(i)$. Formally, we construct the $\alpha(i)$ for each $i$ in the following three steps. First, we add undirected edges to the graph by neighboring relations: if $i$ or $j$ is one of $K$ nearest neighbors of the other, then we add an undirected edge $(i, j)$ to the graph. Second, we assign each edge a direction from the smaller index to the larger one, e.g. the edge $(i, j)$ points from $j$ to $i$ if $j < i$. Finally, we make adjustments to make sure that each $i$ has $\min(K, i - 1)$ parents. If a data point gets inadequate or exceeding parents from the last step, then we add more neighbors to or remove exceeding parents from its parent set. For the latter case, the parents with large indices would be removed first. 

Below we show that $q(\boldf)$ is able to match any covariance values corresponding to graph edges. This result indicates that the order of data points is not important as long as the graph has edges (in either direction) between near neighbors. In our empirical experience, there are only neglectable performance differences with 
permutations of data points.

{\bf Theorem: } Suppose $\bV^*$ is the covariance matrix of the true posterior. By setting appropriate $\bL$, the matrix $\bV = \bL \bL^\top$ is able to match $\bV^*$ at entries corresponding to graph edges, that is, $\bV_{ij} = \bV^*_{ij}, \forall ~~i, j \in \alpha(i)$. 


\subsection{Optimization of the ELBO}
In this section, we optimize the ELBO with stochastic optimization, which needs a cheap unbiased estimation of the objective and its gradient. The strategy is first decomposing the ELBO in \eqref{elbo}, into small factors and then 
estimating its value based on a random selection of factors. Based on this unbiased estimation, we calculate stochastic 
gradients with respect to $\bmu$ and $\bL$. For convenience of discussion below, 
we call the three terms in \eqref{elbo} as the expected likelihood term $L_{ell} = \E{q(\boldf)}{p(\by | \boldf)}$, 
the cross entropy term $L_{cross} = \E{q(\boldf)}{p(\boldf)}$, and the entropy term $L_{ent} = \entropy{q}$. 
Now we decompose the three terms as follows.  

\parhead{Decompose likelihood term} Observations $\by$ are independent given $\boldf$, 
so the likelihood term natually decomposes as in \eqref{eq:hatlell}. We only need a random batch $S$ of data points to 
estimate the entire term $L_{ell}$. We further approximate the expectation term $\E{q(f_i)}{\log p(y_i | f_i)}$ 
for data point $i \in S$ with a Monte Carlo sample $\hat{f}_i$ from $q(f_i)$.
We use reparameterization technique \citep{kingma13}: the random sample $\hat{f}_i$ is a function  
of $\bL_i$, and the gradient of $\tilde{L}_{ell}$ with respect $\bL_i$ is propagated through $\hat{f}_i$. 
The exact calculation and its estimation is shown in \eqref{eq:hatlell}.
\begin{align}
L_{ell} &= \sum_{i=1}^N \E{q(f_i)}{\log p(y_i | f_i)}, \nonumber \\ 
\tilde{L}_{ell} &= \frac{N}{|S|}\sum_{i \in S} \log p(y_i | \hat{f}_i). \label{eq:hatlell}
\end{align}

\parhead{Decompose the entropy term} The entropy term is the determinant of $\bV$ plus a constant. We have $\det(\bV) = \prod_{i=1}^N \bL_{i,i}^2$ by the decomposition of $\bV$. Then the exact entropy calculation and its estimation with a random batch $S$ is
\begin{align}
    L_{ent} &= 0.5(N\log(2\pi e)  + \sum_{i=1}^N \log{\bL_{i,i}^2}), \nonumber \\
\tilde{L}_{ent} &= 0.5(N\log(2\pi e) + \frac{N}{|S|} \sum_{i \in S} \log{\bL_{i,i}^2})~.
\end{align}

\parhead{Decompose the cross entropy term} The cross entropy term is 
\begin{multline}
L_{cross} = - 0.5 \log \det(\bSigma) - 0.5 N \log(2\pi)  \\
- 0.5 ~ \tr{\bSigma^{-1}(\bL\bL^\top + \bmu\bmu^\top)}.
\end{multline}
Let's neglect the term $\log \det(\bSigma)$, which is constant with respect to $q(\boldf)$, and focus on the trace term. The inverse of $\bSigma$ is notoriously hard to compute, so people often appeal to approximations. 

Nystr\"om method \citep{rasmussen06} approximates the kernel matrix with a low-dimensional decomposition. It is not hard to get a low-dimensional decomposition of the precision matrix, and suppose it is $\bSigma^{-1} \approx \bR\bR^{\top}$, then the approximation is 
\begin{multline}
\tilde{L}_{cross} \propto - \frac{1}{2} \frac{N}{|S|} \sum_{i \in S} \big(N_i' (\bL_i\bL_{j_i'}^\top + \mu_i \mu_{j_i'}) (\bR_i\bR_{j_i'}^\top) \\
+ \sum_{j \in \alpha(i) \cup \{i\}}  (\bL_i\bL_{j}^\top + \mu_i \mu_j) (\bR_i\bR_{j}^\top)\big). \label{eq:nystr_cross}
\end{multline}
Here $j_i'$ is a random data point not in $\alpha(i)$, and $N_i' = N - |\alpha(i)| - 1$ . The second term in the bracket is to improve the efficiency of the estimation, as $\bL_i\bL_{j_i'} = 0$ for many $(i, j_i')$ pairs when $i$ and $j_i'$ does not share parents. 
FITC approximation \citep{naish-guzman2008} can be treated likely.

If we use the approximation in \citep{datta16},  
then the prior distribution approximately decomposes as $p(\boldf)\approx \prod_{i=1}^{N}p(f_{i} | f_{\alpha(i)})$ by the graph $G$ we used in the posterior. The conditional $p(f_i | f_{\alpha(i)})$ is derived from the GP prior and has mean 
and variance 
\begin{multline}
    \eta_{i|\alpha(i)} = \bb_i^\top f_{\alpha(i)}, ~\bSigma_{i|\alpha(i)} = \bSigma_{ii} - \bSigma_{i, \alpha(i)} \bb_i^\top \\
    \mbox{with}~~~ \bb_i = \bSigma_{i,\alpha(i)} \bSigma_{\alpha(i), \alpha(i)}^{-1} \label{bandsigma}~.
\end{multline}
Then the prior is decomposible, and an unbiased estimation of the cross entropy term is 
\begin{align}
   \tilde{L}_{cross} =& \frac{N}{|S|} \sum_{i \in S} \E{q(f_i, f_{\alpha(i)})}{\log p(f_i|f_{\alpha(i)})} \label{eq:cond_cross}\\
=&\frac{N}{|S|} \sum_{i \in S} -\frac{1}{2}
\bSigma_{i|\alpha(i)}^{-1} \Big(\mathbf{Q}\mathbf{Q}^\top 
+ (\mu_i-\bb_i^\top \mu_{\alpha(i)})^2\Big) \nonumber \\
& + const, \nonumber
\end{align}
where $\mathbf{Q} = \bL_i-\bb_i^\top \bL_{\alpha(i)}$. We get an unbiased estimation of the ELBO by putting together estimations of the three terms.  
\begin{eqnarray}
L \approx \tilde{L}_{ell} + \tilde{L}_{cross} + \tilde{L}_{ent}. \label{eq:est_elbo}
\end{eqnarray}
The approximation $\tilde{L}_{cross}$ can be either \eqref{eq:nystr_cross} or \eqref{eq:cond_cross}. The second one is neater as it uses a prior with same structure as the variational distribution. We use the second one in our experiments. We then run stochastic optimization to maximize the ELBO. 

This is our first inference method, and we call it NPVI. For a single data point $i$, the most expensive calculation is \eqref{eq:cond_cross}, which takes time $O(K^2)$. For a batch $S$ of $N_b$ data points, the calculation in \eqref{eq:est_elbo} only takes time $O(N_b K^2)$. The gradient calculation takes a similar amount of time. Prior the run of our algorithm, we need to check nearest neighbors and construct the graph -- the time complexity is $O(n\log n)$ with advanced data structures such as k-d trees. 

\subsection{Inference Network to Reduce Model Parameters} 

The number of parameters in the inference method above is in the same scale as the data size. Though each training batch is fast, it needs many iterations to converge when the data size is large. In this subsection, we train an inference network to directly recognize variational parameters $\mu_i$ and $\bL_{i}$ from observations and the prior. 

We define an inference network $\mathcal{M}$ such that, $(\mu_i, \bL_i) = \mathcal{M}(\by_{(i, \alpha(i))}, \bSigma_{(i, \alpha(i)), (i, \alpha(i))})$. Ideally, all observations $\by$ and the entire covariances $\bSigma$ should be in the input, as they all affect $\mu_i$ and $\bL_i$ for data point $i$. For the sake of feasibility, here we only use a small part of related data as the input and hope that the network can still reach good values for the two parameters without complete information. Note that the network only needs to output non-zero elements of $\bL_i$. 

We train two separate Graph Convolutional Networks (GCN) \citep{kipf2017} for $\mu_i$ and $\bL_i$, as the inference network $\mathcal{M}$. A GCN computes hidden layers with the adjacency matrix of a graph and takes input from graph node values. Here we treat data points ${i} \cup \alpha(i)$ as a small graph with $\bSigma_{(i, \alpha(i)), (i, \alpha(i))}$ as the adjacency matrix and $\by_{(i, \alpha(i))}$ as values at graph nodes. In this graph, $i$ is unique because we are computing parameters for $i$. To break the symmetry relation between $i$ and any other data point in $\alpha(i)$, we set $\bA = [\bSigma_{i, i}, 0; \bSigma_{\alpha(i), i}, \bSigma_{\alpha(i), \alpha(i)}]$ as the adjacency matrix for the GCN. The input to the GCN is $\by_{(i, \alpha(i))}$. In the GCN, the hidden layer $\bH^{(l+1)}$ is computed from a previous layer $\bH^{(l)}$ as follows,   
\begin{eqnarray}
    \bH^{(l+1)} = \sigma(\bD^{-\frac{1}{2}} \bA \bD^{-\frac{1}{2}}\bH^{(l)}\bW^{(l)}).
\end{eqnarray}
Here $\bD$ is a diagonal matrix, whose diagonal is the row sum of $\bA$. $\bW^{(l)}$ is the trainable weights for the $l^{th}$ layer. $\bH^{(0)}=\by_{(i, \alpha(i))}$. The activation function $\sigma(\cdot)$ is set to be identity for the last layer and ReLU for other layers. The variational mean $\mu_i$ is obtained by taking the mean of the last layer. 

Reuse of this network for different data points greatly reduces the number of optimization parameters. After learning from several iterations of inferences, it will directly identify good values for variational parameters in later iterations. Therefore, it dramatically reduces the number of iterations for convergence. The GCN learning process is especially suitable for the inference task here: GCN uses the adjacency matrix to consider correlations with nearby observations, which is exactly needed in the inference procedure. 

We call this inference method as NPVI-NN. The complexity of each batch is similar to NPVI. The only extra calculation is the neural network, which takes time  $O(K^2)$ if we treat the size of the neural network as constant.

\section{Experiment}
We evaluate our two methods on three tasks, bird abundency estimation, raster data modeling, and flight delay regression. We put the details of the three tasks in their respective   subsections.  

We compared our methods with three state-of-the-art methods, SVGP~\citep{hensman15},  SAVIGP~\citep{dezfouli15}, and DGP~\citep{cheng17}, for general GP inference. All three methods follow the scheme of inducing points. We use  the implementation of SVGP by GPFlow~\citep{GPflow2017}, the implementations of SVAIGP by its authors, and the implementation of DGP by \citet{faust2018}. We implement noise distribution for these methods when necessary.   

We use the RBF kernel for all experiments. The length scale is a hyper-parameter, which is selected from the candidate set $\{0.05, 0.1, 0.5, 1.0, 1.5, 2.0, 2.5\}$ by checking performance on a validation set.  We randomly split each dataset into three subsets for training (70\%), validation (10\%), and testing (20\%). We report negative log-likelihood on the test dataset as the performance measure for the five algorithms in comparison. The maximum training time for each algorithm on the training dataset is 13.89 hours (50ks).

SVGP, DGP, NPVI, and NPVI-NN are all optimized by stochastic optimization with mini-batches. The optimizer for these methods is Tensorflow AdaGrad. The learning rates of NPVI is set to 0.2 while the learning rates of SVGP, DGP, and NPVI-NN are set to 0.1, based on their respective convergence speeds on the validation set. The size of each mini-batch is 50 for the four methods. SAVIGP uses L-BFGS-B optimizer with default settings from the package. We test SVGP and SAVIGP with $M=200, 1000, 2000$ random inducing points. With any number over 2000, SVGP and SAVIGP will take very long time to converge.  DGP uses separate inducing points to estimate the variational mean and covariance. The paper suggests to use less number of inducing points for covariance estimation. We use 200 inducing points for covariance estimation and vary the number of inducing points ($M=200, 1000, 2000$) for mean estimation. We also have tried $M=4000$ for DGP and found only insignificant performance improvement. We collect the result when the algorithm converges on the validation set or reaches the time limit. For NPVI and NPVI-NN, we vary $K$ as $K=10, 20, 40$. We have tested different GCN structures for NPVI-NN and found typical configurations of the network can give reasonably good performances. We finally decide a network structure with three hidden layers with dimensions 20, 10, and 1.

\begin{table*}[t]
\centering
\caption{Negative log-likelihood of comparison methods on eBird data. Smaller values are better.}
\label{tab:exp4-ebird}
\scalebox{0.90}{\begin{tabular}{ |c|c|c|c||c|c|c| } 
 \hline
$M$ & SVGP & SAVIGP & DGP & $K$ & NPVI & NPVI-NN \\
 \hline
 200  & 2.29$\pm$.10 \textbackslash 691s
 \quad& 2.26$\pm$.07 \textbackslash 220s 
 \quad& 2.05$\pm$.06 \textbackslash 201s
 & 10 & \textbf{1.57$\pm$.04} \textbackslash 23ks& 1.78$\pm$.05 \textbackslash 5.1s \\ 
 \hline                                                                               
 1000 & 2.28$\pm$.10 \textbackslash 50ks ~& 2.20$\pm$.07 \textbackslash 939s ~& 1.99$\pm$.06 \textbackslash 212s &
 20 & \textbf{1.61$\pm$.04} \textbackslash 15ks& 1.79$\pm$.05 \textbackslash 10s~~\\ 
 \hline                                                                               
 2000 & 2.38$\pm$.10 \textbackslash 50ks ~& 2.20$\pm$.07 \textbackslash 2.7ks & 1.98$\pm$.06 \textbackslash 256s
 & 40 & \textbf{1.60$\pm$.04} \textbackslash 50ks & 1.79$\pm$.05 \textbackslash 36s~~\\
\hline
\end{tabular}}
\end{table*}


\begin{figure*}[t]
\begin{subfigure}{.5\textwidth}
\centering
\scalebox{0.9}{
\begin{tabular}{|c|c|c|c|} 
 \hline
 M    & SVGP & SAVIGP & DGP\\ \hline
 200  & 0.5  & 1.0    & 1.5\\ \hline                                        
 1000 & 0.5  & 0.5    & 1.0\\ \hline                                        
 2000 & 0.5  & 0.5    & 1.0\\ \hline
 K   & NPVI & NPVI-NN & \multicolumn{1}{}{}\\ \cline{1-3}
 10  &  0.1 & 0.1     & \multicolumn{1}{}{}\\ \cline{1-3}
 20  &  0.1 & 0.1     & \multicolumn{1}{}{}\\ \cline{1-3}
 40  &  0.1 & 0.1     & \multicolumn{1}{}{}\\ \cline{1-3}
\cline{1-3}
\end{tabular}}
\caption{}
\end{subfigure}
\begin{subfigure}{.3\textwidth}
\centering
\includegraphics[width=1.25\textwidth,trim={0 0 0 0cm}, clip]{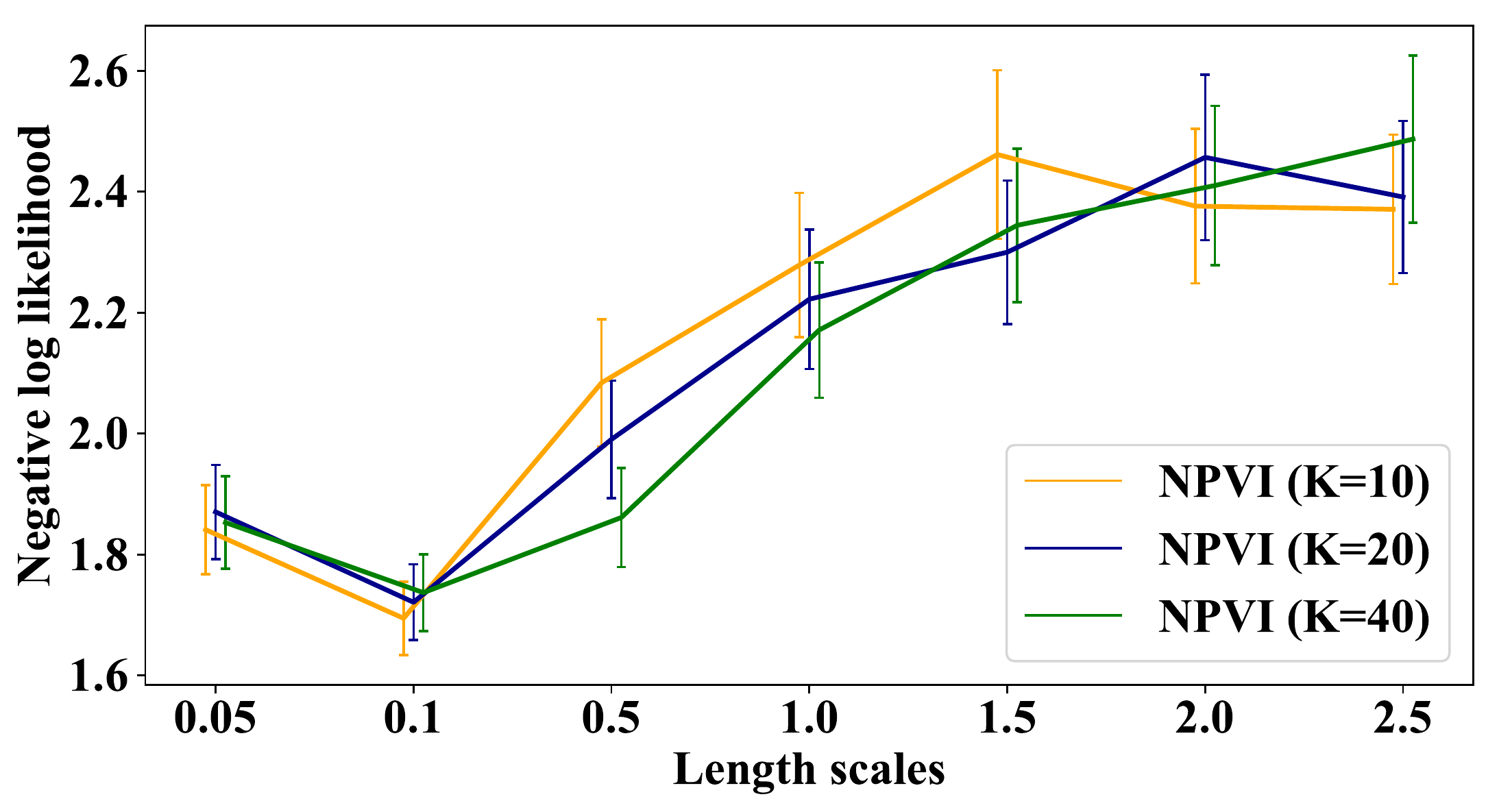}
\caption{}
\end{subfigure}
\caption{
(a) Different length scale chosen through cross-validation by SVGP, SAVIGP, DGP, NPVI, and NPVI-NN with different inference parameters.
(b) Negative log likelihood values with different length scales on validation set.
}
\label{fig:exp1-line-table}
\end{figure*}

\begin{figure*}[t]
\centering
\begin{subfigure}{.38\textwidth}
  \includegraphics[width=1.0\linewidth]{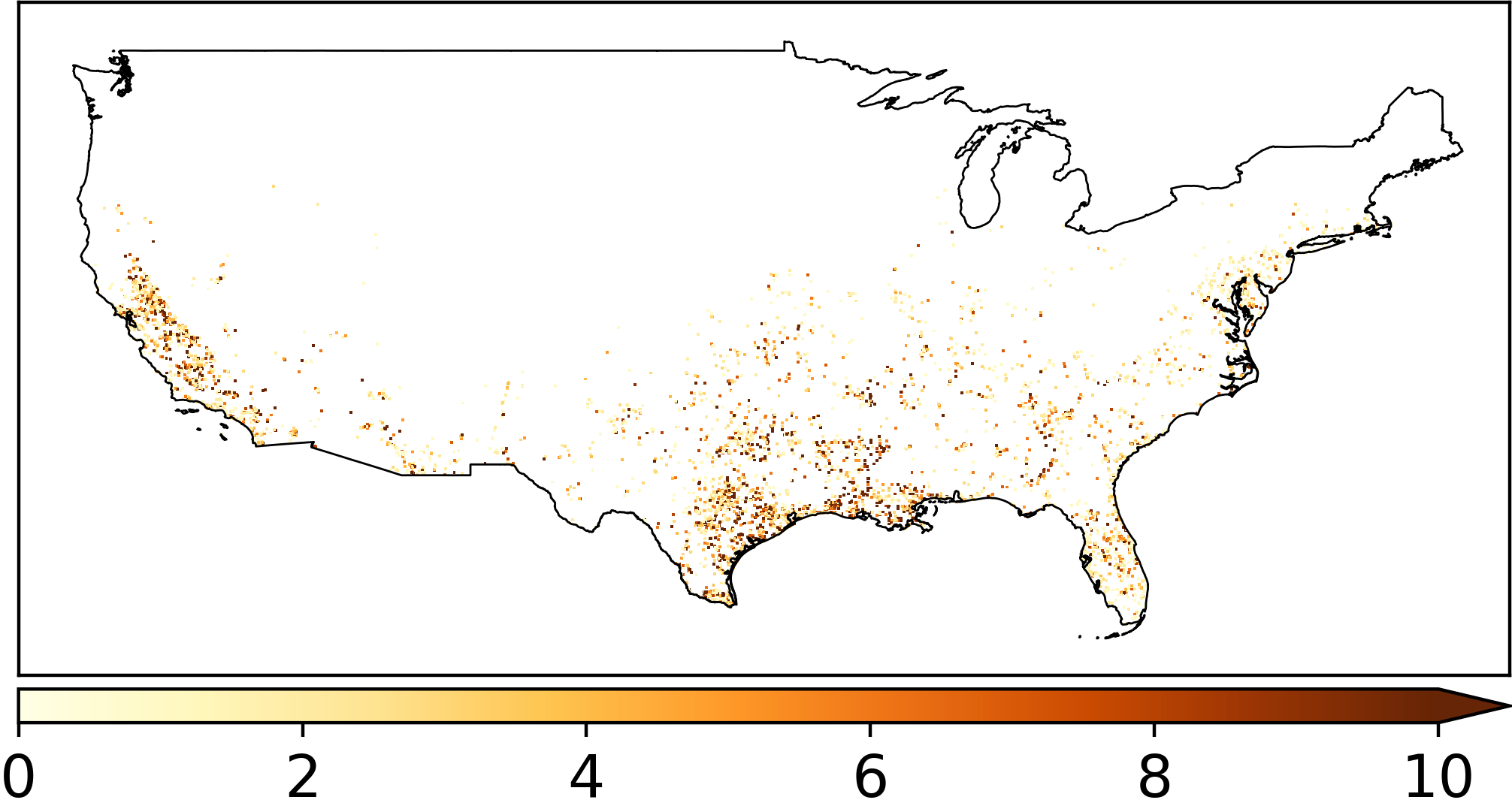}
  \caption{Oberseved data}
  \label{fig:exp3-ebird-true}
\end{subfigure}%
\hskip 1cm
\begin{subfigure}{.38\textwidth}
  \includegraphics[width=1.0\linewidth]{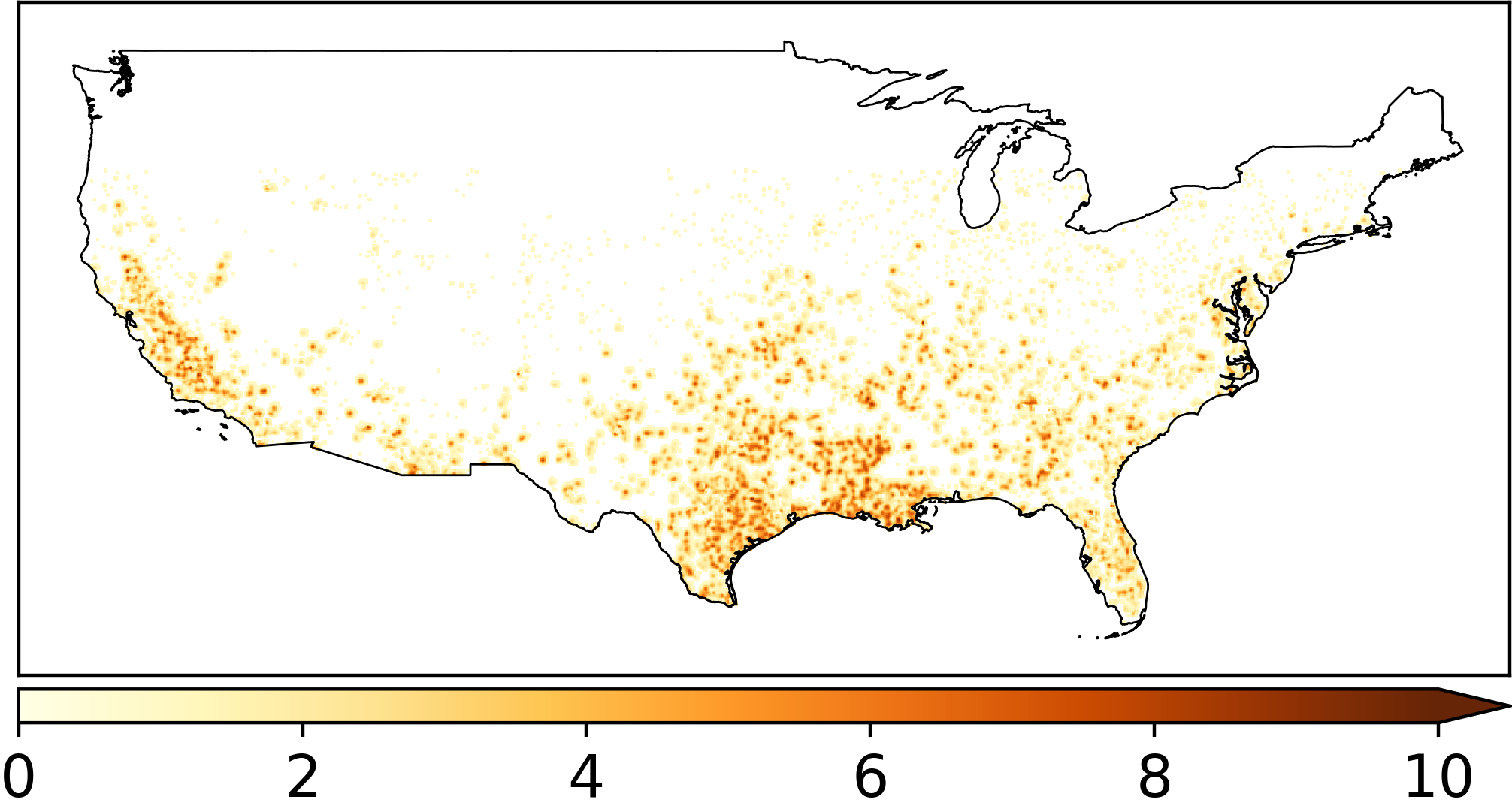}
  \caption{NPVI}
  \label{fig:exp3-ebird-SNNGP}
\end{subfigure}

\begin{subfigure}{.38\textwidth}
  \centering
  \includegraphics[width=1.0\linewidth]{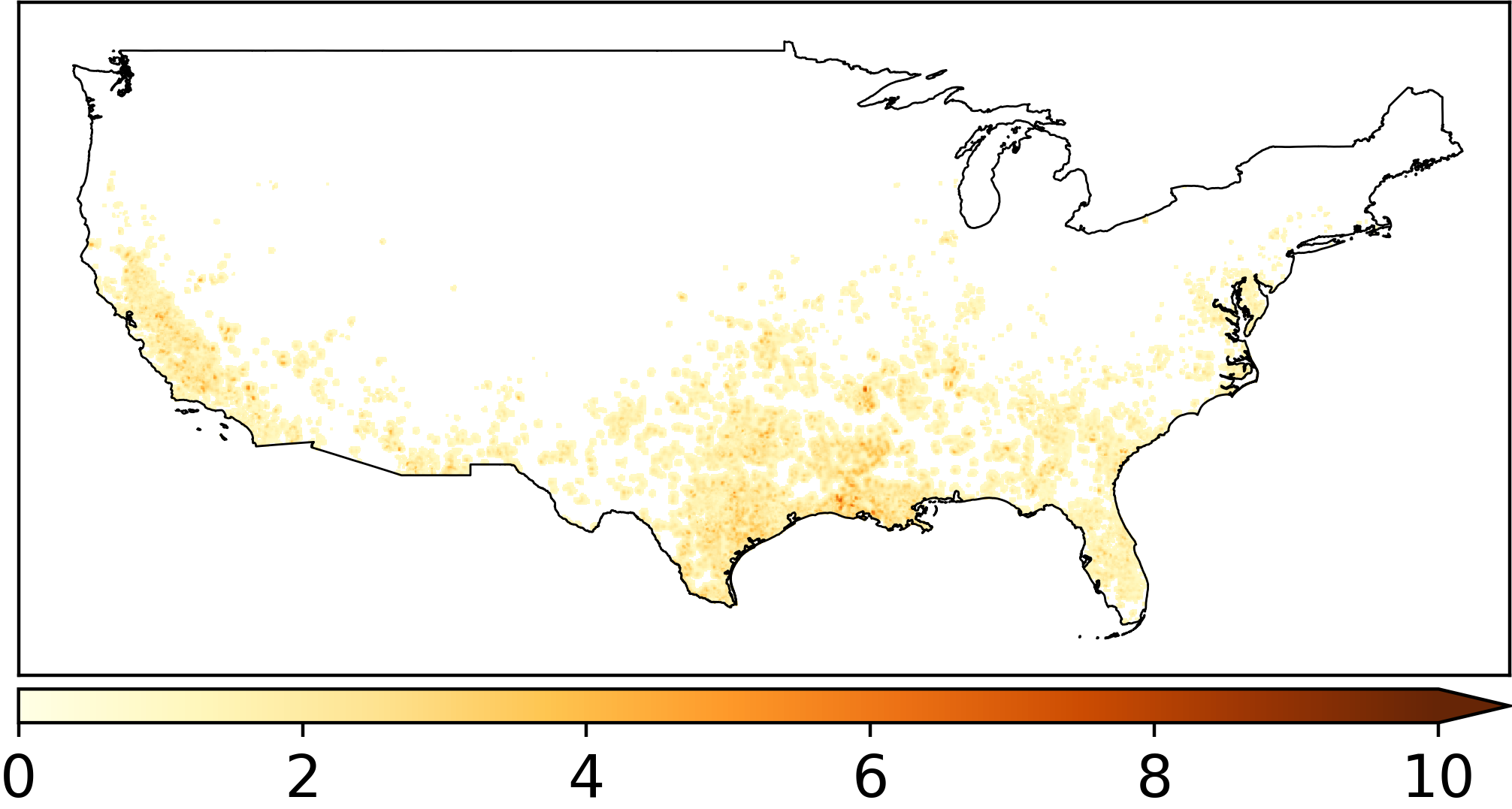}
  \caption{NPVI-NN}
  \label{fig:exp3-ebird-NPVI-NN}
\end{subfigure}
\hskip 1cm
\begin{subfigure}{.38\textwidth}
  \centering
  \includegraphics[width=1.0\linewidth]{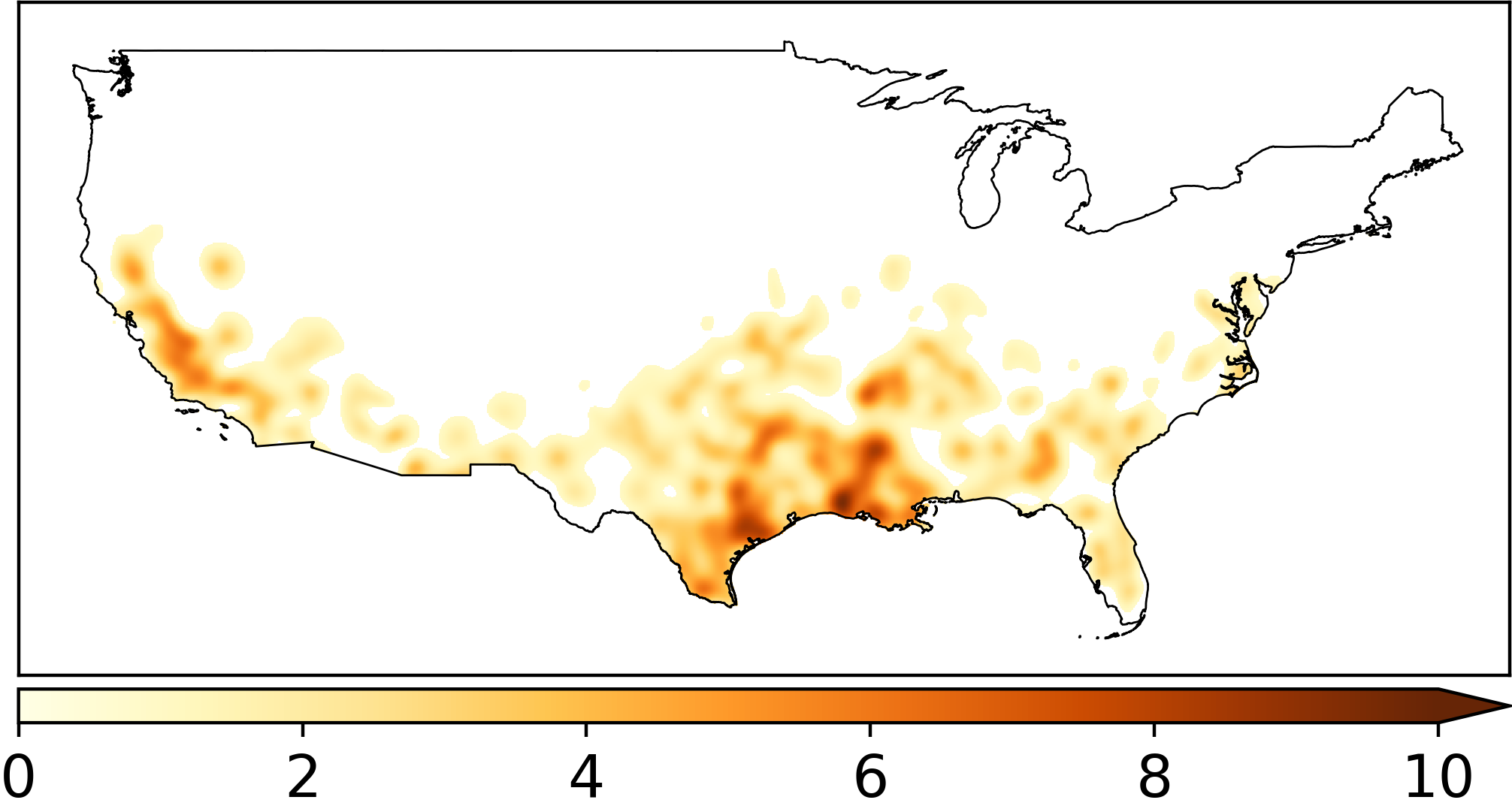}
  \caption{SVGP}
  \label{fig:exp3-ebird-SVGP}
\end{subfigure}
\caption{eBird dataset for the distribution of savannah sparrow in February between 2012 and 2016 across the US. (a) the observed distribution, (b)-(d) the regression surface of the NPVI, NPVI-NN, and SVGP, respectively.}
\label{fig:exp3-ebird}
\end{figure*}

\parhead{The eBird data}
The eBird project \citep{sullivan2009} hosts a repository for the bird-watching community to report bird observations at different locations around the world. In this experiment, we consider the population of the bird species {\it Savannah Sparrow} in February from 2012 to 2016 across the contiguous United States. In our data preprocessing step, we retain all positive observations and a random selection  
from exceedingly many zero observations -- the dataset after preprocessing has 15,085 observations with 6,477 positive data points. We also rescale the input location values such that an Euclidean distance of $0.1$ corresponds to a geographic distance of 20 miles. The link function in \eqref{gp} is defined as the softplus function, $\lambda(x) = \log(1+\exp(x))$. The distribution of an observation $y_i$ at location $i$ is Poisson with mean $\lambda(f_i)$.

Table~\ref{tab:exp4-ebird} shows the negative predictive log-likelihood on the test set. Smaller values mean
better predictive performances. We observe that NPVI and NPVI-NN significantly outperform the other three approaches. The standard deviations obtained from NPVI and NPVI-NN are also the smaller than the SVGP, SAVIGP, and DGP. This result also indicates that NPVI and NPVI-NN is able to achieve good inference performance with a small number $K$. 


We also investigate how the inference method impacts the choice of hyper-parameters. 
Figure \ref{fig:exp1-line-table}(a) shows the length scale chosen by different methods through cross validation. The results indicate that SVGP, SAVIGP, and DGP prefer a smooth prior. As we have analyzed before, a small length scale weakens the correlation between data points and inducing points, then the three methods with a too small number of inducing points will get poor inference results. With this limitation, SVGP, SAVIGP, and DGP have to choose large length scale values. The trend of length scale values chosen by SVGP, SAVIGP, and DGP in \ref{fig:exp1-line-table}(a) is one evidence supporting our argument here. The trend also indicates an even smaller length scale should be used if there are more inducing points.

Figure \ref{fig:exp1-line-table}(b) shows validation perforamnce of NPVI with different length scale and $K$ values. NPVI consistently prefers length scale value $0.1$. Combining Figure \ref{fig:exp1-line-table} (a) and (b), we hypothesize that the length scale 0.1 chosen by NPVI is ideal for the application. This choice limit the influence of a data point within the distance of 50 miles, which is also reasonable in ecology.

In Figure~\ref{fig:exp3-ebird}, we plot the data (a) and the surface predicted by the NPVI (b), NPVI-NN (c), and SVGP (d). For visualization purpose, we put use large pixels to indicate data points. The surfaces inferred by NPVI and NPVI-NN keep more detailed information, which is good for prediction considering its performance in Table \ref{tab:exp4-ebird}. 

\begin{figure*}[t!]
\centering
\begin{minipage}{.4\textwidth}
\centering
\includegraphics[width=0.8\textwidth]{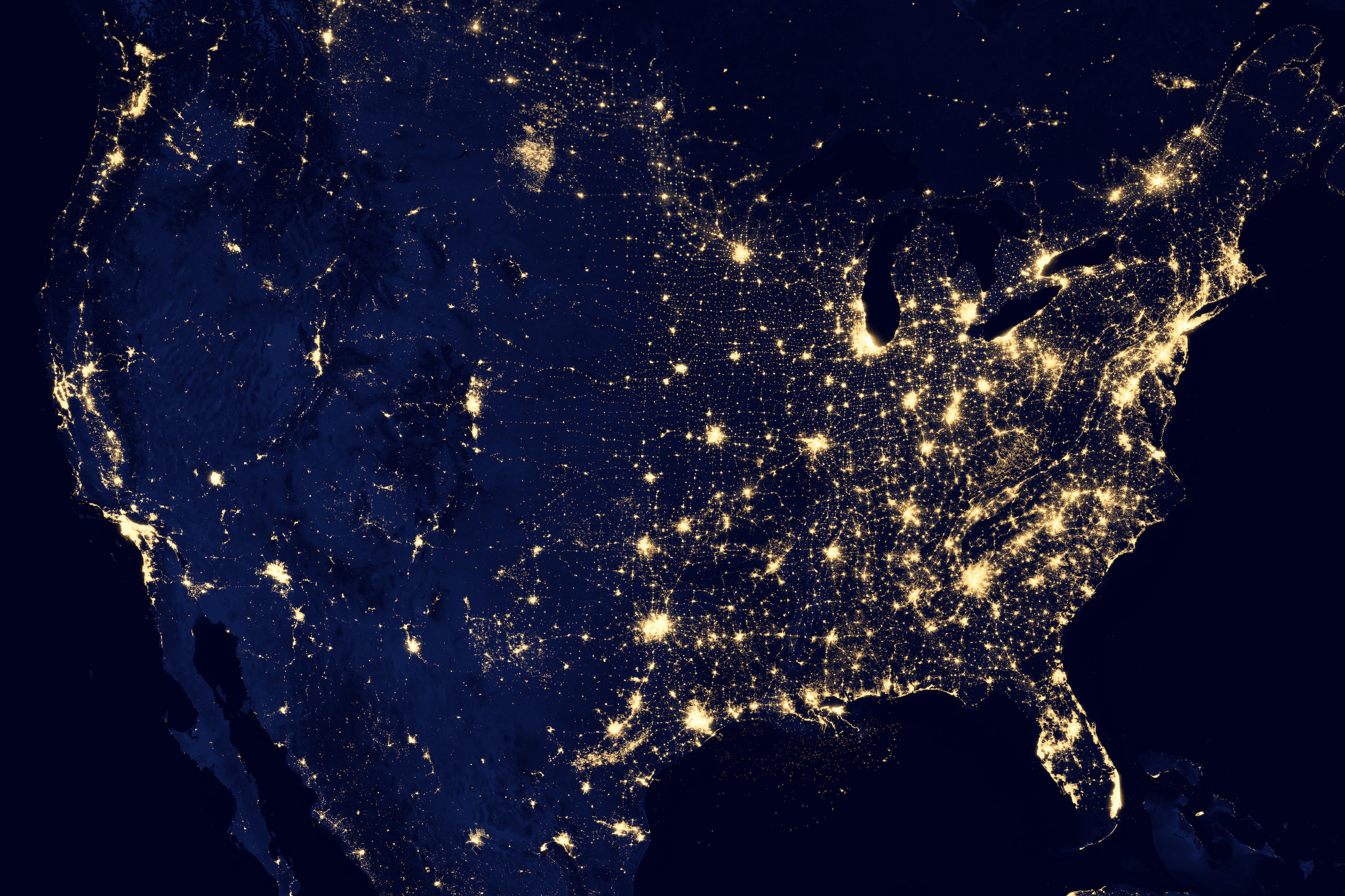}
\caption{The raster data of night time light.}
\label{fig:night-light}
\end{minipage}
\hskip 0.1cm
\begin{minipage}{.505\textwidth}
\centering
\includegraphics[width=0.8\textwidth]{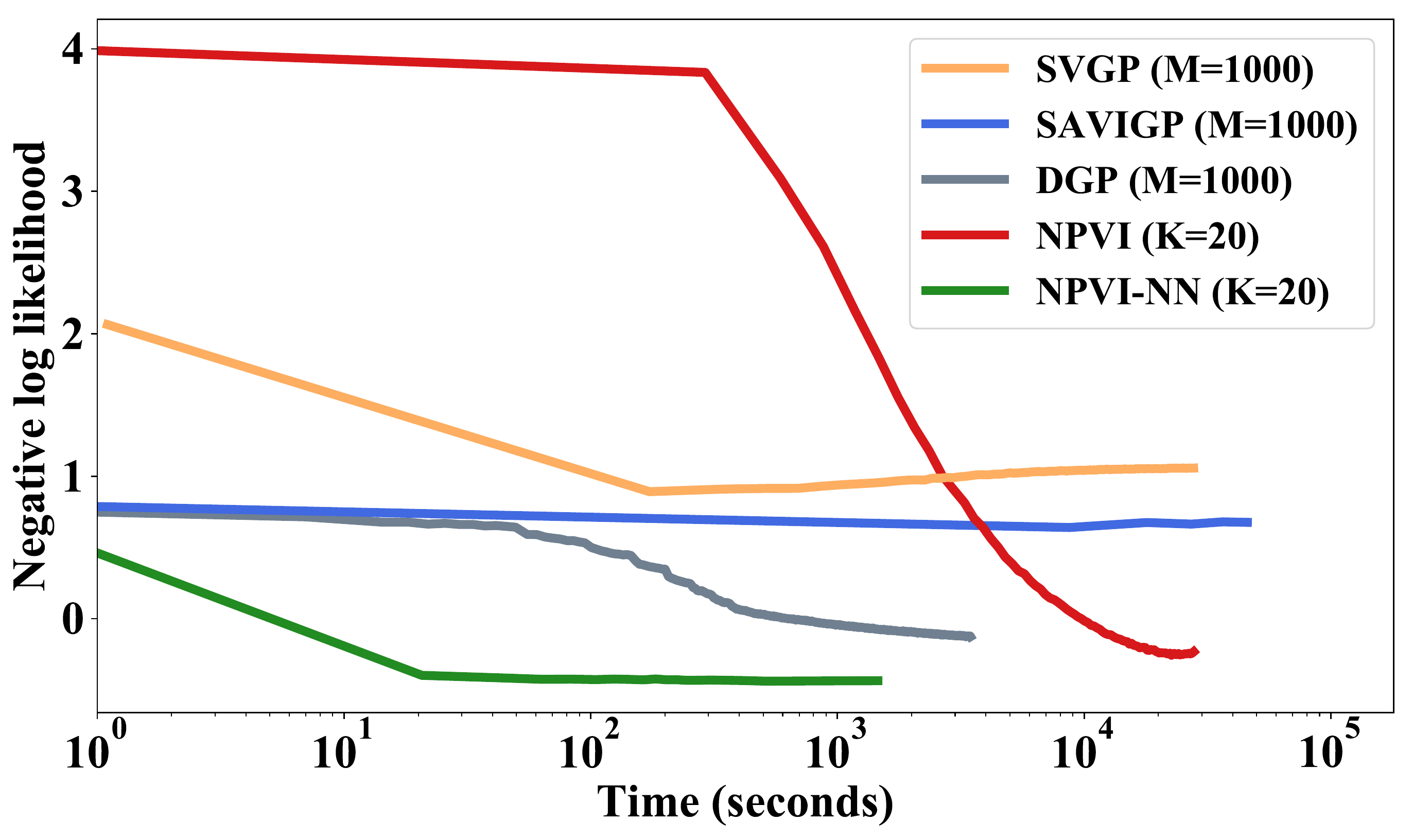}
\caption{Training time of different methods. Y axis is the negative log-likelihood on validation data.}
\label{fig:exp2}
\end{minipage}%
\end{figure*}

\parhead{The NightLight raster data}
In this task, we fit spatial raster data, the NightLight image, with GP models. The data is a satellite image of the night time light over US contiguous states by \citet{NASAData} project (Figure~\ref{fig:night-light}). We use GP models to fit the pixel values over the image. In this experiment, we preprocess the image by first converting the RGB color to greyscale and then downscaling the image. Finally, we get 56,722 pixels with their values rescaled to the range [0, 1]. We use a log-normal distribution with variance parameter $\sigma^2=0.01$ as the noise distribution \dist (the actual standard deviation of the distribution is between 0.1 and 0.27). The mean parameter of the noise distribution $\dist$ is the Gaussian output $f$.

Table~\ref{tab:exp4-light} shows the negative predictive log-likelihood of all methods. NPVI and NPVI-NN both outperform the other three methods. On this dataset, NPVI-NN performs even better than NPVI. It is because NPVI does not fully converge within the time limit. NPVI-NN often converges in the first training epoch.   

We compare the running speed of all methods. Figure~\ref{fig:exp2} shows the negative log-likelihood values in validation against training time for all five methods. There is not an equivalent setting for our methods and a setting for inducing-point methods, so we just choose typical settings for both type of methods. NPVI outperforms SVGP and SAVIGP within reasonable time. NPVI-NN converges much faster than all other methods.

\begin{table*}[t!]
\caption{Negative predictive log-likelihood on NightLight data. Smaller values are better.}
\label{tab:exp4-light}
\centering
\scalebox{0.90}{\begin{tabular}{ |c|c|c|c||c|c|c| } 
\hline
$M$ & SVGP & SAVIGP & DGP & $K$ & NPVI & NPVI-NN\\
 \hline
 200   & 0.71$\pm$.04 \textbackslash 674s & 0.72$\pm$.01 \textbackslash 867s & ~ 0.45$\pm$.01 \textbackslash 1.1ks &
 10 & -0.36$\pm$.01 \textbackslash 22ks & \textbf{-0.47$\pm$.01} \textbackslash 17s\\ 
 \hline                                                                                
 1000  & 0.34$\pm$.03 \textbackslash 50ks\quad& 0.64$\pm$.02 \textbackslash 11ks & -0.12$\pm$.01 \textbackslash 1.3ks &
 20 & -0.20$\pm$.03 \textbackslash 50ks&  \textbf{-0.42$\pm$.01} \textbackslash 27s\\ 
 \hline                                                                                
 2000  & 0.41$\pm$.03 \textbackslash 50ks & 0.64$\pm$.03 \textbackslash 16ks & -0.24$\pm$.01 \textbackslash 1.5ks &
 40 & -0.07$\pm$.02 \textbackslash 50ks& \textbf{-0.46$\pm$.01} \textbackslash 93s\\
 \hline
 \end{tabular}}
\end{table*}

\begin{table*}[t!]
\caption{Negative predictive log-likelihood on Flight data. Smaller values are better.}
\label{tab:exp4-flight}
\centering
\scalebox{0.92}{\begin{tabular}{ |c|c|c|c||c|c| } 
\hline
$M$ & SVGP           & SAVIGP  & DGP & $K$ &  NPVI-NN \\
 \hline                                                                                  
200  & -1.07$\pm$.04  \textbackslash 43s~~~ & -0.90$\pm$.06 \textbackslash 13ks & -1.10$\pm$.002 \textbackslash 770s~ &
10 & \textbf{-1.27$\pm$.002} \textbackslash 34s~~~\\ 
 \hline                                                                                  
1000 & -1.01$\pm$.03 \textbackslash 1.1ks & -0.84$\pm$.04 \textbackslash 50ks & -1.04$\pm$.002 \textbackslash 900s~ & 
20 &\textbf{-1.28$\pm$.002} \textbackslash 86s~~~\\ 
 \hline                                                                                  
2000 & -1.02$\pm$.05 \textbackslash 3.8ks & -0.86$\pm$.07 \textbackslash 50ks & -1.11$\pm$.002 \textbackslash 3.3ks & 
40 & \textbf{-1.27$\pm$.002} \textbackslash 533s\\ 
\hline
\end{tabular}}
\end{table*}


\parhead{Flight Delay dataset}
This data set contains 5.9 million records of flight arrivals and departures within the US in 2008\footnote{From http://stat-computing.org/dataexpo/2009/}.
Following \citet{hensman13}, we select 8 dimensions as the attributes and randomly choose 800,000 instances from the original dataset as the dataset for this experiment. The task is to use the 8 attributes to fit the delay time. In the data pre-processing step,  we standardize input features, remove extreme delay time values, and then translate and scale of all delay time to [0, 1]. We set the noise distribution $\dist$ to be the log-normal distribution with $\sigma^2=0.01$ and $\mu$ being the GP output.

The predictive performances and training time of the four methods are shown in Table~\ref{tab:exp4-flight}. We can see that NPVI-NN has the best predictive performance and also runs much faster than the other three methods. We observe that NPVI-NN converges after training with about 100k data points (about one sixth epoch). NPVI cannot run on this dataset due to too many parameters.

\section{Conclusion}
In this work, we propose two new variational inference methods, NPVI and NPVI-NN, for GP models with general noises. The two methods has  a highly decomposible ELBO , so they can run stochastic optimization efficiently to learn the variational distribution. NPVI and NPVI-NN retains the flavor of non-parametric learning by inferring function values directly at data points instead of through inducing points. 
NPVI-NN reduces the number of variational parameters by training an inference network to recognize variational parameters from nearby data points. It only uses a relatively small number of iterations to converge and greatly speeds up the inference procedure.

\nocite{langley00}

\bibliography{sections/spatial.bib}
\bibliographystyle{icml2018}


\appendix
\setcounter{table}{0}
\renewcommand{\thetable}{A.\arabic{table}}
\begin{table*}[!ht]
\caption{Negative predictive log-likelihood on Precipitation data. Smaller values are better.}
\label{tab:exp4-precipitation}
\centering
\scalebox{0.82}{\begin{tabular}{ |c|c|c|c||c|c|c| } 
\hline
$M$ & SVGP & SAVIGP & DGP & $K$ & NPVI & NPVI-NN\\
 \hline
 200   & -0.64$\pm$.17 \textbackslash 1.6ks & -0.29$\pm$.20 \textbackslash 364s~ & -1.26$\pm$.02 \textbackslash 917s ~&
 10 &  \textbf{-1.57$\pm$.03} \textbackslash 37ks & -1.48$\pm$.03 \textbackslash 30s \\ 
 \hline                                                                                
 1000  & -1.24$\pm$.12 \textbackslash 34ks \quad& -0.76$\pm$.20 \textbackslash 1.5ks & -1.30$\pm$.02 \textbackslash 1.0ks &
 20 & \textbf{-1.60$\pm$.02} \textbackslash 20ks &  -1.42$\pm$.02 \textbackslash 50s \\ 
 \hline                                                                                
 2000  & -1.26$\pm$.16 \textbackslash 50ks~ & -1.12$\pm$.16 \textbackslash 3.1ks & -1.31$\pm$.02 \textbackslash 1.2ks &
 40 & \textbf{-1.55$\pm$.02} \textbackslash 22ks &~  -1.33$\pm$.02 \textbackslash 179s \\
 \hline
 \end{tabular}}
\end{table*}
\section*{Appendix}
\section{Theorem Proof}
{\bf Theorem: } Suppose $\bV^*$ is the covariance matrix of the true posterior. By setting appropriate $\bL$, the matrix $\bV = \bL \bL^\top$ is able to match $\bV^*$ at entries corresponding to graph edges, that is, $\bV_{ij} = \bV^*_{ij}, \forall ~~i, j \in \alpha(i)$.\\

{\bf Proof: } Define $\beta(i) = \alpha(i) \cup \{i\}$. The covariance of  $f_{\beta(i)}$  and $f_i$ in the variational distribution $q(\boldf)$ is 
\begin{align}
  \bV_{\beta(i), i} =  \bL_{\beta(i)}\bL_{i}^\top
  =\bL_{\beta(i), \beta(i)}\bL_{i, \beta(i)}^\top.
\end{align}
Note that the row $\bL_i$ has non-zero elements only at $\beta(i)$. 
Because $\bL$ is a lower triangular matrix, its principal sub-matrix $\bL_{\beta(i), \beta(i)}$ is a full-rank matrix. 
Then we just set $\bL_{i, \beta(i)} = (\bL_{\beta(i), \beta(i)}^{-1}\bV^*_{\beta(i), i})^{\top}$ so that 
$\bV_{\beta(i), i} = \bV^*_{\beta(i), i} $. 
\QEDB

\section{Random ordering on eBird dataset}
We run the NPVI-NN 10 times and each time randomly permute the training examples. The standard deviation of negative log-likelihood for $K$=10, 20, 40 are all around 0.02. This results indicate that the data ordering does not have a significant influence during the inference.

\section{The Precipitation data}
In this experiment, we analyze monthly precipitation averaged from 1981 to 2010 across the contiguous United States provided by \citet{NOAAData}. We choose the data in May for this experiment. There are 8,833 locations, each associated with the average precipitation level. In our pre-processing step, we scale precipitation values into the range of $[0, 1]$. The noise distribution is a log-normal distribution with variance parameter $0.01$.The mean parameter of the log-normal distribution is the output of GP $f$.

From the results in Table~\ref{tab:exp4-precipitation}, we can see that NPVI achieves the best performance. The NPVI-NN converges much faster while still getting comparable inference accuracy as NPVI.

\end{document}